\documentclass[runningheads]{llncs}
\usepackage[T1]{fontenc}
\usepackage{graphicx}

\usepackage{tabularx}
\usepackage{comment}
\usepackage{enumerate}
\usepackage[shortlabels]{enumitem}
\usepackage{wrapfig}

\usepackage{briefcite}

\usepackage{ifthen}
\newboolean{briefcitations}
\setboolean{briefcitations}{true}  

\ifthenelse{\boolean{briefcitations}}{
  \includecomment{brief}
  \excludecomment{full}
}{
  \excludecomment{brief}
  \includecomment{full}
}

\begin{document}
\title{A Proposal to Extend the Common Model of Cognition with Metacognition }

\titlerunning{CMC Metacognition}

\author{John Laird, Christian Lebiere, Paul Rosenbloom, Andrea Stocco}
\institute{Contact: John.Laird@cic.iqmri.org}
\authorrunning{Laird, Lebiere, Rosenbloom, Stocco}

\maketitle              
\begin{abstract}

The Common Model of Cognition (CMC) provides an abstract characterization of the structure and processing required by a cognitive architecture for human-like minds. We propose a unified approach to integrating metacognition within the CMC. We propose that metacognition involves reasoning over explicit representations of an agent's cognitive capabilities and processes in working memory. Our proposal exploits the existing cognitive capabilities of the CMC, making minimal extensions in the structure and information available within working memory. We provide examples of metacognition within our proposal. 

\end{abstract}

\section{Introduction}
The Common Model of Cognition (CMC) \cite{laird_standard_2017} 
was developed as an abstract consensus model of \textit{human-like minds}, derived from the computational structures and representations of cognitive architectures \cite{kotseruba_40_2020}, and informed by our knowledge of the human mind and brain. The CMC specifies the fixed \textit{architectural} structures for encoding, maintaining, using, and acquiring \textit{knowledge} to produce \textit{behavior}, emphasizing routine cognition. 
Here, we propose extending the CMC to include metacognition \cite{cox_raja_2011,flavell_metacognition_1979,johnson2025,kralik_metacognition_2018,nolte2025,walker_2025}.

Broadly speaking, metacognition encompasses reasoning about \textit{any} aspect of cognition, including reasoning, memory, perception \cite{Rahnev2022}, motor skills, and learning. It can include partial or even incorrect theories of cognition, such as reasoning about perceived ESP capabilities. We focus on metacognition related to an agent's own cognition. We also discuss \textit{metareasoning}: reasoning about reasoning as a restricted form of metacognition. Examples of metacognition include: introspective monitoring, deliberate decision making, deliberate learning \cite{laird_learning_2018}, predictive and hypothetical reasoning, retrospective reasoning, strategy selection, and self-explanation. 

\begin{figure}[t]
\centerline{\includegraphics[width=1\linewidth]{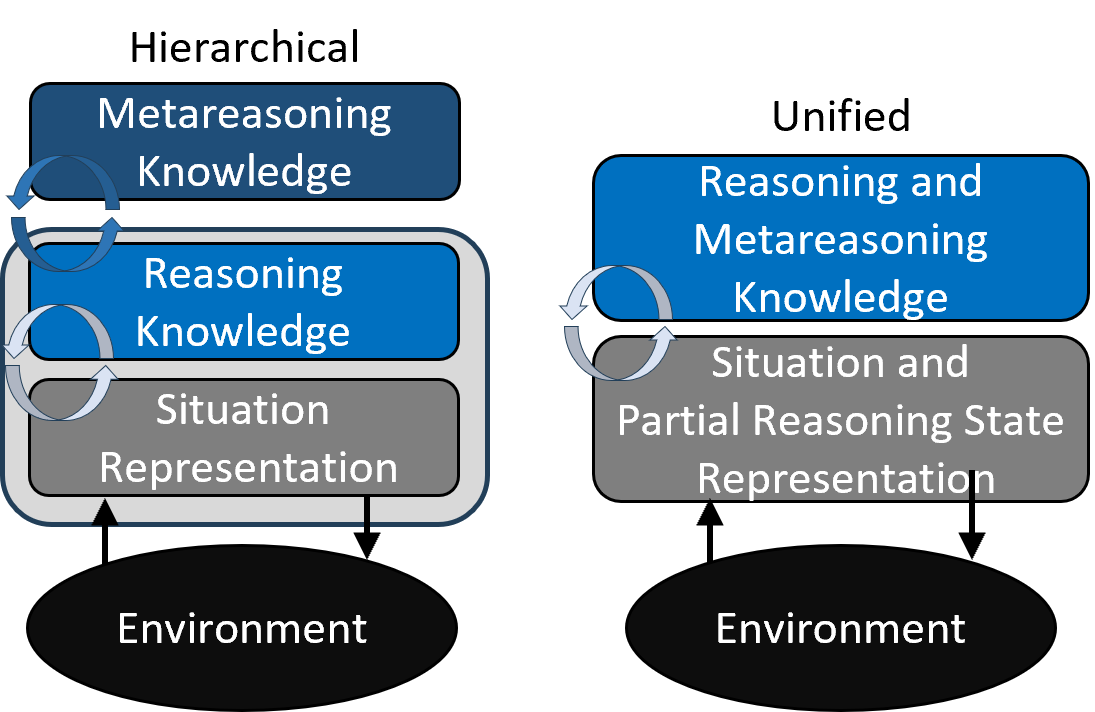}}
\caption{Alternative Metacognitive Architectures.}
\label{MCArch}
\end{figure}

Figure \ref{MCArch} shows two approaches to metacognition in cognitive architectures. In the hierarchical approach, also referred to as the Nelson and Narens model \cite{nelson_narens_1990}, specialized modules are added ``on top'' of cognition to implement metacognition, as exemplified by MIDCA \cite{cox_midca_2016} and Clarion \cite{Sun01062007}. Those modules monitor cognition, reason about it, and modify its processing. They access the current state of cognitive processing, the histories of processing, and representations of the agent's procedural knowledge that drives cognition, with reasoning and metareasoning operating in parallel without intermixing. 

In the CMC and cognitive architectures more generally, a capability is realized through architectural structures and knowledge. Therefore, we propose a \textit{unified} approach, on the right side of Figure \ref{MCArch}, where cognition and metacognition differ only in what is the subject of reasoning. We propose minimal architectural extensions to make information \textit{about} an agent's cognition available in working memory. Versions of these extensions are found in existing CMC architectures, including ACT-R \cite{anderson_integrated_2004},  Sigma \cite{rosenbloom_sigma_2016}, and Soar \cite{laird_soar_2012}. We also describe the sources of other non-architectural sources of information about cognition needed to support metacognition as a component of overall cognition. As with the original goals for the CMC, our goal is for our proposal of metacognition to apply to both humans and A(G)I systems with similar general capabilities.

Our proposal does not permit information in long-term memories to be examined or reasoned over by other modules. Instead, it restricts reasoning and metareasoning to information available in working memory. It avoids new modules by incorporating the long-term knowledge used in metareasoning within its existing long-term memories. Restricting access to long-term memories sacrifices omniscient metareasoning but enables efficient processing and memory functionalities consistent with neural memory models. Furthermore, all existing cognitive reasoning and learning capabilities are available for metacognition, including interaction with the external environment. 

Below, we review the Common Model of Cognition and present our proposal for extending it to include human-like metacognition. Our proposal focuses on adding new representational distinctions and sources of information \textit{about} cognition that become available to an agent to initiate, reason, and terminate metacognition. We attempt to identify the minimum architectural information required to support general metacognition, as well as non-architectural sources of information about cognition that are available to an agent. We include three examples of metacognition within this framework. 

\section{Common Model of Cognition}

\begin{figure}[t]
\centerline{\includegraphics[width=1\linewidth]{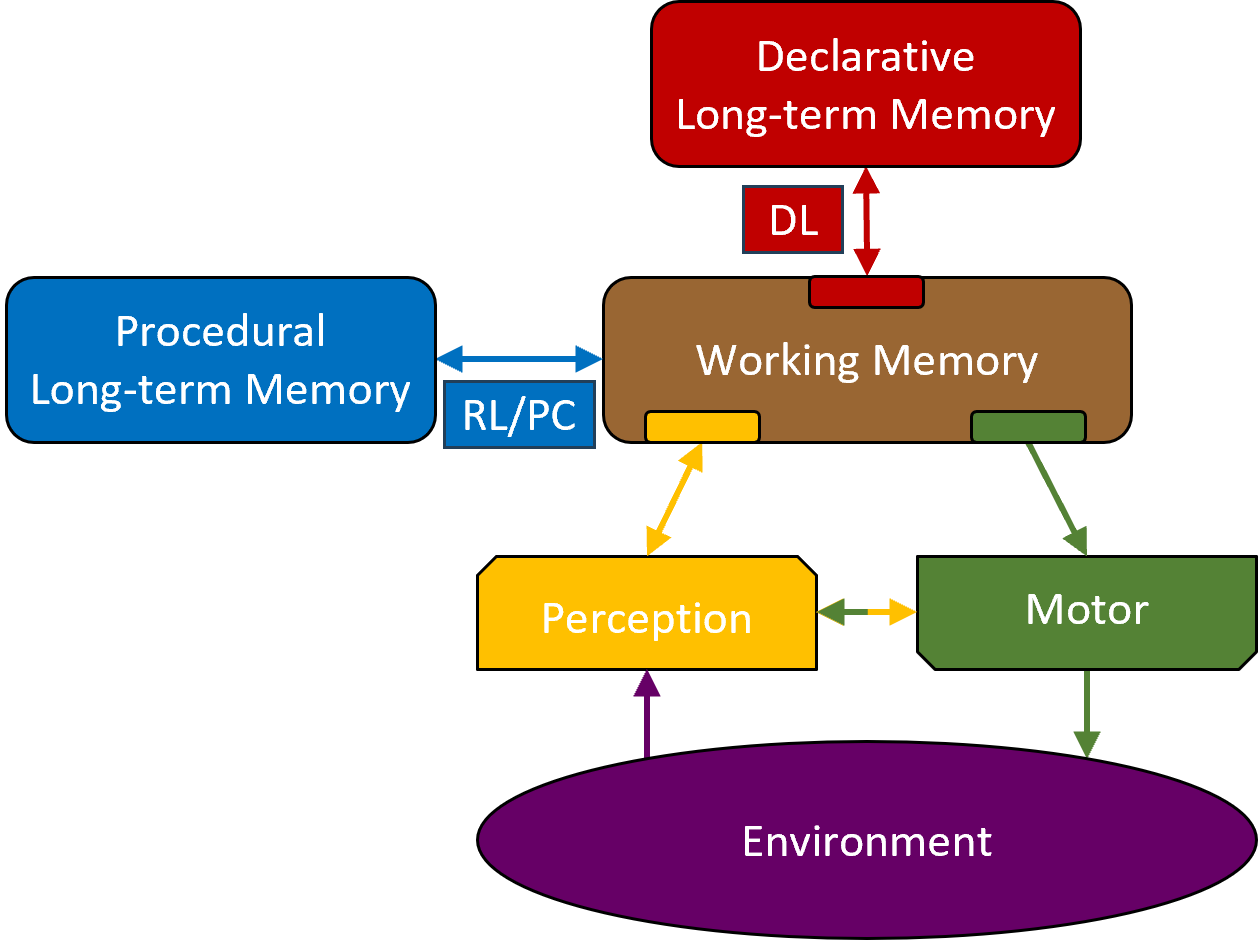}}
\caption{The Common Model of Cognition.}
\label{CMCFigure}
\end{figure}
The CMC unifies many similar cognitive architectures by identifying common components, processing, connectivity, and constraints. It does not specify mechanisms or implementations, but emphasizes functionality, focusing on routine cognition and learning. Cognition is the collective processing of the component modules (working memory, procedural memory, declarative memory) and their associated processes (action selection, retrieval, learning).

Figure \ref{CMCFigure} shows the structure and data flow among the modules, which include short-term working memory, long-term procedural memory, long-term declarative memory, learning, perception, and motor control. Long-term memories have associated automatic learning mechanisms that incrementally modify and extend their contents. The CMC posits procedural compilation, reinforcement learning, and declarative learning. Memories contain relations over symbols annotated with \textit{quantitative metadata}. Examples include the recency of creation or access, probability, and derived utility. Metadata influences processing within a module, such as retrieval and learning.

Data flow begins with perception and proceeds to working memory, representing an agent's understanding of its current situation and goals. Procedural memory contains knowledge about selecting and executing actions that modify working memory. On each \textit{cognitive cycle,} procedural memory, testing the contents of working memory, selects a \textit{single} action, which makes one or more changes to working memory, resulting in a \textit{step} in the cognitive cycle. Each of the other modules has a buffer in working memory through which procedural memory can initiate memory retrievals, motor actions, or top-down control of perception. Results from those module processes are added to their respective buffers. Thus, behavior unfolds as a sequence of steps, driven by procedural memory, making changes to working memory. We call such step-by-step behavior \textit{reasoning} where the contents of working memory are what the reasoning is \textit{about} and the knowledge in procedural and declarative memory determines the course of the reasoning. The original CMC encompasses routine and skilled performance, where working memory includes only information about the agent's current goals and task state.

\section{Phases of Metacognition Processing}
In our proposal, metacognition employs the same cognitive cycle process, utilizing the same modules, and often the same knowledge. Metacognition is distinguished by the fact that the information about the agent's own processing is included in its reasoning: information about prior problem solving (``I see where I made a mistake on that problem and need to do something different in the future.''), specific skills (``I'm good at jigsaw puzzles.''), general competences (``I struggle with math.''), and the operation of individual modules (``I have trouble remembering the difference between \textit{affect} and \textit{effect}.''). Our proposal does not introduce new modules but relies on existing architectural structures and knowledge, which are extended to make new forms and representations of information available in working memory. The process of metacognition is not purely metacognition, but a combination of new and existing capabilities. 

In this section, we describe the three phases of metacognition: initiation, reasoning, and termination. We identify the types of knowledge needed in these phases. Then we describe our proposed extensions to meet those requirements, as well as sources of knowledge available for metareasoning. The following section steps through those phases using three examples of metacognition. 

\subsection{Initiation}
The standard cognitive cycle for reasoning involves procedural memory responding to changes in working memory, typically concerning the performance of the current task - what we call base-level reasoning. However, a working memory element can be added that indicates a deviation of base-level reasoning, such as a failure in the retrieval from memory. The general requirement is that a knowledge source creates a structure in working memory that is \textit{about} the agent's cognition. Such an item can be created deliberately through an action of procedural memory, but also as a side-effect of other long-term memory retrievals or perception. Our proposal extends these by including feedback about the state of individual modules. 

\subsection{Reasoning}
Once information is available about the state of the agent's cognition, the agent can use its cognitive capabilities to respond to it (or ignore it), essentially treating it as the signal that a new (meta)problem may need to be solved. Knowledge for reasoning can combine existing task performance knowledge with knowledge that is specific to metacognition. We have identified three extensions to the existing sources of information that can be crucial for metacognition. 
\begin{enumerate}
    \item Information about the current state of agent processing in its modules. 
    \item A memory of the contents of working memory over time, where a sequence of past situations can be retrieved into working memory and reasoned over. This allows the agent to detect choices it should not make in the future or ones it should reinforce. It also allows the agent to learn a model of the effects of its actions on its internal state, its environment (changes that come from perception), and its processing state. 
    \item Some means of creating past or future hypothetical states in working memory with two seemingly contradictory properties. One is that these states are represented such that the agent's knowledge for reasoning in the current state can apply to them, so that the agent can imagine what it would do in those states. The second is that they are also distinguished in some way, so that the agent does not confuse them with reality. Thus, an agent can plan for the future or retrospectively reconsider past actions, without disrupting or interfering with its reasoning about the present.
\end{enumerate}

\subsection{Termination}
Metacognition terminates when reasoning is no longer sensitive to representations of the agent's processing, such as when metacognition resolves the reason it was initiated. Depending on the initiating signal, this can be from a deliberate change to working memory, or indirectly because an impasse in reasoning is resolved. Whatever the reason for termination, any result can also change long-term memory (via the existing learning mechanisms).  

\section{Proposal for Metacognition in the CMC}
The crux of our proposal is to expand the sources and representations of an agent's available information to meet the requirements described above. We present our proposal in two stages. First, we describe structural modifications to the CMC that make new representations and \textit{direct} sources of information available. In the second, we describe how the existing sources can also provide some of that information \textit{indirectly}. 

\begin{figure}[t]
\centerline{\includegraphics[width=1\linewidth]{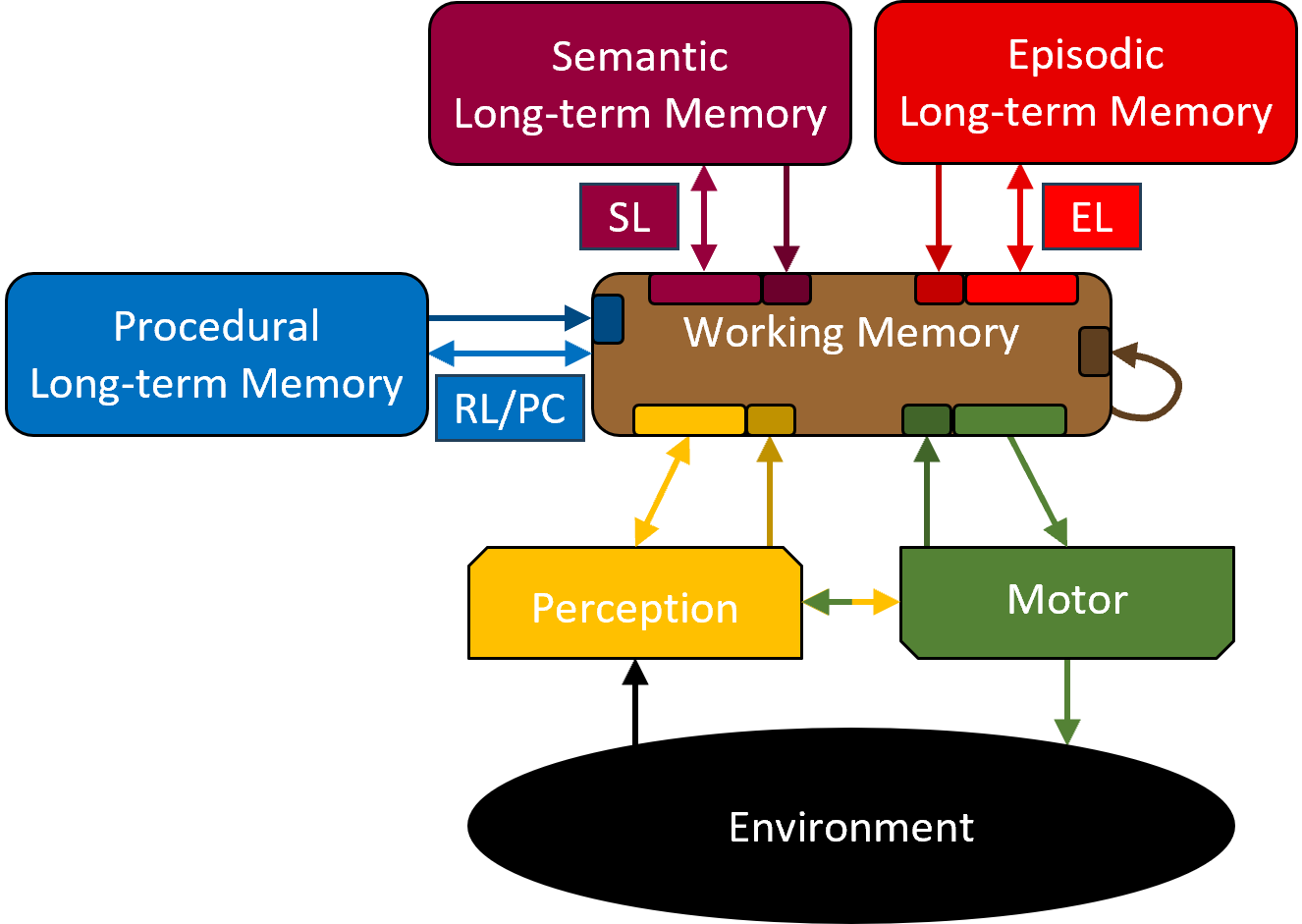}}
\caption{Structural extensions to the CMC to support Metacognition.}
\label{CMC2Figure}
\end{figure}

\subsection{CMC Extensions}
Figure \ref{CMC2Figure} illustrates our proposed extensions. One is that information about the current processing state of each module, including procedural memory and working memory, is available in working memory buffers. The second is the inclusion of the functionality of episodic memory, which is shown as a separation from semantic memory. The third (not shown) is that working memory supports representations of information about past or future states, distinguished from the current state. We also describe how each extension is, or is not, implemented in ACT-R, Sigma, and Soar. 

\subsubsection{Module Process-state Buffers} 
In our proposal, each module has a \textit{process-state} buffer added to working memory. It summarizes information about the module's state that can be a signal to initiate metareasoning, as well as the information for understanding the current state of processing during metareasoning. Types of process-state information include the success or failure of requested actions for a given buffer. Additional information can include certainty/confidence in a result, partial results (the answer starts with an 'A'), or indications that an answer is available but not retrieved (feeling of knowing). Perceptual process-state information can include surprise or the inability to recognize parts of the perceptual scene. Process-state information for working memory can include assessments, such as desirability or intrinsic pleasantness. A proposed CMC extension for emotion \cite{rosenbloom_proposal_2024} considers adding a metacognitive assessment module with an associated buffer in working memory, which would be compatible with this proposal. 

The existing module buffers in ACT-R, Sigma, and Soar include dedicated areas for process-state information. Sigma and Soar include process-state information associated with procedural memory, whereas ACT-R does not. In Sigma and Soar, if procedural memory cannot select an action, a structure called the \textit{substate} is added to working memory. The substate describes the reason for the impasse and initiates metareasoning and provides a context for metareasoning (see below). Sigma, but not ACT-R or Soar, has something akin to the working memory process-state buffer for representing desirability appraisals.  

\subsubsection{Hypothetical State Representations} 
Our proposal requires that an agent can currently represent hypothetical past or future states with the current state in working memory. The substate mechanisms in Sigma and Soar provide concurrent representations of the current situation for base-level reasoning and hypothetical states and are the locus for metareasoning. The structure of the substates can recreate the current state so that the agent's long-term knowledge can be used for both metareasoning and base-level reasoning. However, the substates are distinguished by certain features so that hallucination is avoided. ACT-R has no similar architectural mechanisms. Instead, in agents that employ planning, knowledge-based conventions reserve specific slots in their working memory structures (chunks) to distinguish hypothetical states. 

\subsubsection{Episodic Memory} 
Process-state buffers provide information about the instantaneous state of the agent, but do not provide any representation of the history of reasoning that can be used for metareasoning. The ability to reconstruct prior reasoning trajectories is precisely what episodic memory can provide. The original CMC had a single long-term declarative memory. Here, we propose that the functionality associated with episodic memory differs from that of semantic memory, as it provides a direct source of knowledge about an agent's past reasoning. Episodic learning incrementally and automatically acquires episodes and their temporal relations, allowing an agent to reconstruct the extended sequence of reasoning in working memory, possibly through multiple retrievals. Once in working memory, that data, which is about the agent's reasoning, enables retrospective analysis and other forms of metareasoning. 

Soar has distinct semantic and episodic memories as shown in Figure \ref{CMC2Figure}; however, ACT-R and Sigma do not. Instead, they have a single long-term declarative memory, where some of the functionalities of episodic memory are supported but not all.  
 
\subsection{Indirect Sources of Processing Information}
In addition to the direct sources provided by process-state buffers and episodic memory, an agent can access information about its processing from its interactions with its environment, by metareasoning, from its memories of its reasoning, behavior, and metareasoning. 

\subsubsection{Perception of its Environment:}
An agent's environment includes many sources of information that an agent can use to reason about itself. Below are a few general categories. 
\begin{itemize}
    \item Self Observation: 
    By observing the results of its interactions with its world, an agent can learn about the impact of its cognitive processes on its environment. When unexpected results occur, the perception of those results can lead to metacognition. When self-observation is combined with episodic memory, an agent can later review its behavior, determining what approaches work and which don't work, and build up a model for future tasks and associated metareasoning. 
    
    \item Other Agents: Other agents can provide observations about an agent's cognitive processing, suggest possible reasoning steps or strategies, notify an agent of mistakes (or successes) in its reasoning, or even provide an agent with cognitive strategies (such as through Cognitive Behavior Therapy) to detect its reasoning approaches, evaluate them, and possibly modify them.  

    \item Recorded Information: 
     An agent can read books, watch movies, and even study psychology to acquire a general understanding of reasoning capabilities that it applies to itself.
\end{itemize}
Many of these sources provide temporal distance between an agent's original reasoning and using the information, allowing the agent to create internal representations of behavior and capabilities in working memory after the behavior is generated. This can be impossible during routine reasoning, when only task information is available and task urgency prevents metacognitive processing. 

\subsubsection{Metareasoning:}
Metareasoning itself composes these other sources of knowledge about an agent's cognitive capabilities to draw conclusions and generate new insights that fuel future metareasoning. This requires the other originating sources of information, but allows an agent to extend and expand its knowledge by combining information from multiple sources, identifying trends and commonalities, and so on. The knowledge created by metareasoning can then be learned and transferred to long-term memory for future use as described below.

\subsubsection{Semantic and Procedural Memories:}
The automatic learning of procedural and semantic knowledge enables the retention of knowledge about an agent's cognition produced by other sources. They are indirect sources, as they require information to be present in working memory, which, through learning, becomes available for retrieval into working memory and future metacognition. 

\section{Examples of Metacognitive Processing}
We return to the three phases of metacognitive processing with three examples. 

\subsection{Wordle Retrieval}
The agent attempts to retrieve a word from semantic memory using a partial specification of a five-letter word.
\begin{itemize}
    \item Initiation:  The semantic memory process-state buffer includes information that the retrieved word (`Tripe') is uncommon. 
     \item Reasoning: The agent decides it wants to know more ``about what it knows about the word,'' and attempts to retrieve the word from episodic memory, embedding it in the context of it being a Wordle answer. Episodic memory does not retrieve a specific episode of it being a previous answer (possibly because of interference from the hundreds of previous times the agent has played Wordle). Still, the process-state buffer indicates that the word is very familiar. The agent reasons that the combination of being uncommon and familiar indicates it is probably a previous Wordle answer, as it wouldn't come up in any other situation. 
     \item Termination: The agent discards `Tripe' from consideration, which terminates the short bout of metareasoning. The agent returns to generating a potential answer.
\end{itemize}

\subsection{Making a Move in Chess}
The agent is playing a chess game and is far enough into the game that its memorized opening moves are exhausted. Here we describe the mechanisms in Sigma and Soar that support metareasoning in such a situation. 
\begin{itemize}
    \item Initiation:  Procedural memory does not return a single definite action to take. In Sigma and Soar, this leads to the creation of a substate. 
     \item Reasoning: In the substate, the agent decides to explicitly try out the different available moves on internal copies of the current state, generating states that it then compares. 
     \item Termination: It ultimately decides on a specific move, which terminates the metareasoning.  
\end{itemize}
 
 \subsection{Repeated Robot Action}
 A one-armed robot is instructed to store all the leftovers on the counter in the refrigerator. Each time the robot stores an item in the refrigerator, it opens the refrigerator door, fetches the item, places it in the refrigerator, and then closes the refrigerator door. 
 \begin{itemize}
    \item Initiation: After completing the task, the robot's instructor tells it that it needs to improve its performance. This external information triggers procedural knowledge that the robot should do a retrospective analysis of its original performance. 
     \item Reasoning: Using existing procedural knowledge, the robot recalls a trace of its behavior from episodic memory into working memory. It then uses existing procedural knowledge to analyze the trace. It detects that it is repeatedly in exactly the same world state because it closes the refrigerator door as part of one store command, but then immediately opens it for the next. It ``imagines'' being at the end of a store command when there are other items to store, and inhibits the action to close the refrigerator door. Through its procedural learning mechanism, it learns to inhibit that action in similar situations in the future.
     \item Termination:  On completing its internal inhibition, it has additional procedural knowledge that removes from working memory the comment from the instructor, and continues with its normal activities. 
\end{itemize}
 
\section{Conclusion}
We propose that three specific extensions be added to the CMC to support a unified approach to metacognition: module process-state buffers that provide information in working memory on the current state of each of the buffers; episodic memory that provides a means for the agent to recreate its behavior so that it becomes available to reason over; and some means of creating hypothetical situations in working memory that support using base-level reasoning during metareasoning, avoid interference and confusion between them. Our proposal is a framework, but it does not specify in detail the diverse and extensive indirect learned or pre-encoded long-term knowledge needed for an agent to engage in the forms of metacognition found in humans. An essential point of our proposals is to provide the architectural structure that is necessary above and beyond what is encoded as knowledge in the long-term memories. 

Although we introduce some new architectural structures, there are no structural boundaries between base-level (non-meta) reasoning and metareasoning. Within existing architectures consistent with this proposal (ACT-R, Sigma, and Soar), an agent can rapidly switch between base-level reasoning and metareasoning, as the only difference is whether working memory structures are about the agent's cognitive and reasoning capabilities. In addition, an agent's reasoning transitions from base-level task reasoning to metareasoning as it responds to internal impasses, failures, and successes in base-level reasoning, and then back to base-level reasoning as learning leads to routine, impasse-free reasoning. 

\newpage
\bibliographystyle{splncs04}
\bibliography{abc-zotero,format}

\end{document}